\documentclass{IEEEtran}

\usepackage{cite}
\usepackage{amsmath,amssymb,amsfonts}
\usepackage{algorithmic}
\usepackage{graphicx}
\usepackage{textcomp}

\usepackage{bm}
\usepackage{amsthm}
\usepackage{xcolor}
\usepackage{balance}
\usepackage[normalem]{ulem}

\newtheorem{theorem}{Theorem}

\newtheorem{corollary}{Corollary}
\newtheorem{remark}{Remark}

\newcommand{\note}[1]{#1}

\newcommand{\myvec}[1]{\mathbf{#1}}
\newcommand{\mymatrix}[1]{\mathbf{#1}}

\def\BibTeX{{\rm B\kern-.05em{\sc i\kern-.025em b}\kern-.08em
    T\kern-.1667em\lower.7ex\hbox{E}\kern-.125emX}}
\begin{document}
\title{\bf \huge
Unifying Complementarity Constraints and Control Barrier Functions for Safe Whole-Body Robot Control
}


\author{Rafael I. Cabral Muchacho$^{1}$, Riddhiman Laha$^{2}$, Florian T. Pokorny$^{1}$,\\ Luis F.C. Figueredo$^{2,3}$, and Nilanjan Chakraborty$^{4}$
\thanks{This work was partially supported by the Wallenberg AI, Autonomous Systems and Software Program (WASP) funded by the Knut and Alice Wallenberg Foundation, and
by StMWi Bayern (Project X, grant no. 5140951).}
\thanks{$^{1}$Division of Robotics, Perception and Learning, KTH Royal Institute of Technology, Sweden. Email: \texttt{ricm@kth.se}, \texttt{fpokorny@kth.se}.
    $^{2}$Munich Institute of Robotics and Machine Intelligence, TUM, Germany. Email: \texttt{riddhiman.laha@tum.de}.
    $^{3}$School of Computer Science, University of Nottingham, UK. Email: \texttt{figueredo@ieee.org}.
    $^{4}$Department of Mechanical Engineering, Stony Brook University, NY, USA. Email: \texttt{nilanjan.chakraborty@stonybrook.edu}
    }}

\maketitle

\begin{abstract}
Safety‑critical whole‑body robot control demands reactive methods that ensure collision avoidance in real-time.   
Complementarity constraints and control barrier functions (CBF) have emerged as core tools for ensuring such safety constraints,
and each represents a well‑developed field.
Despite addressing similar problems, their connection remains largely unexplored. 
This paper bridges this gap by 
formally proving the equivalence between these two methodologies for sampled-data, first-order systems, considering both single and multiple constraint scenarios.  
By demonstrating this equivalence, we provide a unified perspective on these techniques. This unification has theoretical and practical implications, facilitating the cross-application of robustness guarantees and algorithmic improvements between complementarity and CBF frameworks. We discuss these synergistic benefits and motivate future work in the comparison of the methods in more general cases.
\end{abstract}

\begin{IEEEkeywords}
Autonomous Robots, Optimal control, Constrained control, Robotics.
\end{IEEEkeywords}

\section{Introduction}
\label{sec:intro}


Safety guarantees in whole-body robot control often require 
strict avoidance of collisions and constraint violations, 
including distance thresholds, manipulator constraints, input constraints, and other requirements~\cite{pupa2021safety,laha2023s,lagomarsino2025pro, muchacho2023shared}. 
Two prominent mathematical frameworks that have demonstrated significant efficacy in formalizing safety-critical planning and control with closed-loop dynamics are complementarity-based methods~\cite{chakraborty2009complementarity,sinha2021task, sinha2023oc3, yao2025synthesis} and 
Control Barrier Functions (CBFs)~\cite{ames2016control, ames2019control, breeden2021control}. For reactive behaviors, both formulations ensure constraint satisfaction through an online optimization program.

While both methods address similar aspects of safety and control, their research developments have predominantly evolved in parallel. 
Consequently, important connections between these two frameworks have remained unexplored, including aspects of their underlying mathematical structure and equivalences.  
We aim to uncover the relationship between the methods by studying the case of safe whole-body robot control, in the sense of collision avoidance, adhering to first-order dynamics in sampled data systems.

Our main contribution is a formal analysis and a proof of equivalence between complementarity-based methods and CBFs for whole-body robot control, in the case of sampled-data first-order closed-loop systems.  
In establishing this equivalence, our objective is not only theoretical. Practical benefits include the transmission of algorithmic improvements,
the transfer of established robustness and safety margins, and the cross-application of existing solvers.
Finally, we provide a numerical example for validation, and highlight valuable connections between the methods in both directions.

The remainder of this paper is organized as follows.
After an overview of related work in Section~\ref{sec:rel_work}, Section~\ref{sec:prelim} introduces the notation and the problem of safe whole-body robot control.
Section~\ref{sec:safe_control} describes the complementarity and CBF approaches to safe control as a preparation to their formal comparison.
Our main result is presented in Section~\ref{sec:equivalence}, and validated through a numerical example in Section~\ref{sec:validation}.
Lastly, Section~\ref{sec:conclusion} provides some final conclusions
and directions for future work.
\section{Related Work}
\label{sec:rel_work}

In the context of whole body safe control, it is imperative to construct dynamical systems that are safe by design. In other words, mathematical tools should formally guarantee collision-free motions.
The need for such tools is particularly pressing in dynamic environments, which demand reactive approaches. Design strategies include complementarity-based approaches~\cite{chakraborty2009complementarity, yao2025synthesis}, and CBF methods~\cite{ames2016control, ames2019control}, 
among others~\cite{khatib1986real, rimon1990exact, Aubin1991book}. 
In this paper, we focus on local methods that prioritize  collision avoidance along the entire robot surface over trajectory tracking or goal convergence.

\textbf{Complementarity-based approaches} have long been employed in control theory and robotics to manage non-smooth constraints, such as stiff-contacts, bounds on inputs,
and computational dynamics~\cite{posa2014direct,zhang2023simultaneous,zhang2025simultaneous}.  
From a control perspective, linear complementarity constraints can be viewed as piecewise affine conditions, ensuring safe behaviour at each step by automatically activating or deactivating constraints based on contact states or distance thresholds. 
Concretely, optimization problems subject to complementarity constraints are used to represent the non ``glue-like" fashion (non-penetration) of rigid body contacts~\cite{le2024contact} and can be efficiently solved~\cite{mitsopoulou1987contribution, dirkse1995path,jourdan1998gauss, erleben2013numerical}. 
These methods have, in turn, motivated the usage of Linear Complementarity Problems (LCP) in effective compliant control, e.g., in obstacle avoidance for mobile robots~\cite{sinha2023oc3}, safe whole-body robot control~\cite{yao2025synthesis}, and in motion planning extensions~\cite{chakraborty2009complementarity, sinha2021task}.

\textbf{Control barrier functions} \cite{ames2016control, ames2019control, breeden2021control}, on the other hand, are rooted in Lyapunov-like arguments for forward invariance. 
In special cases, CBFs can be casted as convex quadratic programs that in turn are solved at run time to enforce constraints on states and keep the system trajectories within a safe set.   
Collision-free control through CBFs 
has been widely used in the context of complex closed-loop electromechanical systems~\cite{ames2016control}. 
As far as articulated systems like manipulators are concerned, barrier-type methods have been proposed for safety preservation at the kinematic~\cite{landi2019safety}, as well as the dynamic level~\cite{choudhary2022energy,singletary2022safety}. 
The key takeaway is that a safe set can be defined in the robot configuration space that provides safety guarantees for the resulting collision-free trajectory without continuous re-planning. 
Further, CBF methods have also been applied to tasks involving physical human-robot collaboration~\cite{shi2023dual,shi2023safety}, where the central idea is to design a function that depends also on time, in addition to the system state. 
The authors in~\cite{murtaza2021real,murtaza2022safety} explore a similar idea to handle operational space constraints, including obstacle avoidance as CBF constraints. 
\note{Finally, it is worth highlighting that CBFs are also commonly used as safety filters layered on top of nominal stabilizing controllers, including, e.g., control Lyapunov functions style constructions in which CLF terms encode performance objectives. These safety-filter QPs are routinely augmented with additional affine constraints—such as actuator bounds, saturation, and input limits—see, e.g., \cite{liu2025safetycritical,2025_Deng_TAC_complex_constraints}. Notwithstanding, safety conditions are still enforced through first-order (affine-in-input) inequalities at the sampling instants, and therefore, the resulting  
equivalence guarantees established herein extend to these settings---whenever the additional requirements enter the same convex QP layer as affine constraints. 
}
For a more detailed tutorial-style usage of CBFs as a tool for safe collision avoidance in the context of articulated robots, we refer interested readers to~\cite{ferraguti2022safety}.

\section{Preliminaries} \label{sec:prelim}

We begin by considering the fully actuated system
\begin{align}
    \dot{\myvec{q}} = \myvec{u}, \label{eq:system}
\end{align}
with n-dimensional joint state $\myvec{q}\in \mathcal{Q} \subset \mathbb{R}^n$, and control input $\myvec{u} \in \mathcal{U} \subset \mathbb{R}^n$.
Let state constraints be described by scalar functions $h_j\in \mathcal{C}^1_{loc}$, i.e., with locally Lipschitz first derivatives,
\begin{equation}
  h_j: \mathbb{R}^n \to \mathbb{R}, 
  \quad j = 1,\dots,m,
\end{equation}
each defining the constraint $h_j(\myvec{q})\geq 0$.
The safe set $\Omega$ is given by the zero superlevel-set 
\begin{align}    
  \Omega 
:= \Bigl\{\,\myvec{q}\in\mathcal{Q} : h_j(\myvec{q})\ge0 \text{ for all } j\Bigr\},
\end{align}
where the gradient $\tfrac{\partial}{\partial \myvec{q}} h_j(\myvec{q})$ is locally Lipschitz continuous 
and non-degenerate on the boundary, 
\begin{align}
    \partial\Omega := \{\myvec{q} \in\mathcal{Q} \mid h_j(\myvec{q})=0,\ \exists j\in [1, \dots, m]; \\ 
    h_i(\myvec{q})\geq 0,\ \forall i\in [1, \dots, m] \setminus j
    \}.
\end{align}
We refer to $\myvec{h} \in \mathbb{R}^m$ as the (column) vector of constraint functions.

\subsection{Constraints for Safe Whole-Body Robot Control}

We focus on the task of safe whole-body robot control. Here, the goal is for a robot to follow a motion policy while avoiding collisions between the manipulator's links and obstacles in the environment.
To this end, a constraint function $h_i$ can be defined as the minimum distance between the $i$-th link of the robot and the obstacle set in task space.

Let the point $\myvec{p}_{c,i}$ be the closest point on the link's surface to the corresponding obstacle surface point $\myvec{p}_{o,i}$, then $h_i(\myvec{q}) = \lVert \myvec{p}_{c,i} - \myvec{p}_{o,i}\rVert$.
The map from configuration is given by $\myvec{f}_{c,i}: \mathcal{Q} \to \mathbb{R}^d$,
\begin{align}
    \myvec{p}_{c,i} = \myvec{f}_{c,i}(\myvec{q}) \in \mathbb{R}^d\\
    \dot{\myvec{p}}_{c,i} = \tfrac{\partial} {\partial \myvec{q}}\myvec{f}_{c,i}(\myvec{q}) \, \dot{\myvec{q}} = \mymatrix{J}_{c,i}\dot{\myvec{q}},
\end{align}
where $d$ is the task space dimension and $\mymatrix{J}_{c,i}$ is the contact Jacobian.
Since the gradient of the euclidean distance function coincides with the unit normal $\myvec{n_i}$, we obtain by chain rule
\begin{align}
    \frac{\partial h_i}{\partial \myvec{p}_{c,i}} = \frac{\myvec{p}_{c,i}^T - \myvec{p}_{o,i}^T}{\lVert \myvec{p}_{c,i} - \myvec{p}_{o,i} \rVert} = \myvec{n}_{i}^T,\quad \frac{\partial h_i}{\partial \myvec{q}} = \myvec{n}_{i}^T\mymatrix{J}_{c,i}.
\end{align}

\begin{remark}
The distance function to non-convex shapes is only almost everywhere $\mathcal{C}^1_{loc}$. Although in practice this is not a limitation, a valid theoretical construction can be obtained by representing the environment and the manipulator as a union of balls to arbitrary precision.
Then, constraints $h_i$ can be defined for every pair of balls, and each is ensured to be $\mathcal{C}^1_{loc}$. 
As an alternative to ball-based representations, one can also leverage $\mathcal{C}^1_{loc}$ relaxations of the euclidean distance to non-convex objects~\cite{li2024representing, liu2022regularized, muchacho2025adaptive}.
\end{remark}

The presented whole-body safety constraints can then be used for safe and reactive robot control in fast changing environments through the approaches described below.
\section{Safe Control Approaches} \label{sec:safe_control}

To track a desired end-effector trajectory, a nominal input trajectory is generally given by joint velocities through differential inverse kinematics.
Two approaches used to achieve safe whole-body robot control are (i) complementarity approaches formulated as quadratic programs with linear complementarity constraints (LCQP) and (ii) invariance-focused approaches through control barrier functions.
In this section, we summarize the approaches and reduce them to their primitive forms as a preparation for their comparison.

\subsection{Complementarity-based Approach}

Complementarity constraints are nonlinear and non-convex constraints denoted by
\begin{align}
    0 \leq a \perp b \geq 0,
\end{align}
which represents the constraints
\begin{align}
    0 \leq a\ \land \ ab=0 \ \land \ b \geq 0.
\end{align}
Constraints of this form have been successfully used to represent contact dynamics as in~\cite{kwak1991complementarity, pang1996complementarity, le2024contact} and have inspired methods for safe kino-dynamic and whole-body control of robotic manipulators~\cite{chakraborty2009complementarity,yao2025synthesis}.

The target behavior in whole-body collision avoidance can be represented as a switched system through complementarity constraints, where the velocity compensating for contact is constrained to zero whenever the distance function surpasses a predefined threshold~\cite{yao2025synthesis, sinha2023oc3}. Mathematically, this switching mechanism is described by the complementarity condition,
\begin{align}
    0 \leq \lambda_i \perp h_i - \delta_{\mathrm{LC}, i} \geq 0,
\end{align}
where $i$ indexes a specific pair of link and obstacle and $\delta_{\mathrm{LC}, i}$ denotes a safety threshold. Here, the term $\lambda_i \in \mathbb{R}_{>0}$ acts as a scaling factor for the unit normal vector of motion $\myvec{n}_i \in \mathbb{R}^d$.

The complementarity constraint is evaluated at the next time step using a first order approximation and time step $\tau$, leading to the differential complementarity problem (DCP) formulation
\begin{align}
    h_i({\myvec{q}_{t+\tau}}) \approx h_i({\myvec{q}_t}) + \tau \dot{h}_i(\myvec{q}_t) \\
    0 \leq \lambda_i \perp h_i({\myvec{q}_{t+\tau}}) - \delta_{\mathrm{LC}, i} \geq 0,\quad i = 1,\dots,m \\
    \myvec{0} \leq \bm{\lambda} \perp \myvec{h}({\myvec{q}_{t+\tau}}) - \bm{\delta}_{\mathrm{LC}} \geq \myvec{0}.
\end{align}

Using the previous definitions and $\mymatrix{J}_{c,i}^\dagger \myvec{n}_{i} = (\myvec{n}_{i}^T\mymatrix{J}_{c,i})^\dagger$, where $\dagger$ denotes the Moore--Penrose inverse,
the velocity in configuration space (input) can be parametrized as
\begin{align}
    \myvec{u} &= \myvec{u}_\mathrm{des} + \sum_{i=1}^{m} \mymatrix{J}_{c,i}^\dagger \myvec{n}_{i}\lambda_i.
\end{align}

In order to provide the policy in matrix-vector form,
\note{%
let $\myvec h(\myvec q)\in\mathbb{R}^m$ collect the constraint functions $h_i(\myvec q)$, and let  
$
\myvec a_i(\myvec q) = \frac{\partial  h_i}{\partial \myvec  q}(\myvec  q)\in\mathbb{R}^{1\times n}.   
$
Then, we can define the Jacobian of the constraints with respect to the configuration (with $i$-th row given by $\myvec a_i$) as  
\begin{equation}
\mymatrix A(\myvec{q}) = \tfrac{\partial \myvec h}{\partial \myvec q}( \myvec q)\in \mathbb{R}^{m\times n}.
\end{equation}
%
Note that geometrically, $\myvec a_i(\myvec{q})$ is normal to the level set $h_i(\myvec{q})$. In particular, when $h_i$ is chosen as a signed distance or clearance function, $\myvec a_i(q)=\nabla_{\myvec q} h_i( \myvec q)$ is (up to scaling) the outward normal direction in configuration space at the current state. We then define the row-wise
Moore--Penrose normalization operator acting on $\mymatrix A(\myvec q)$ as 
\begin{equation}
\mymatrix G {:} \left(\mymatrix A \right) \mapsto 
\begin{bmatrix}
    \myvec a_1^\dagger & \cdots & \myvec a_m^\dagger
\end{bmatrix}
\in\mathbb{R}^{n\times m},
\label{eq:g-operator}
\end{equation}
assuming $\|\myvec a_i \|\neq 0$ for all $i$. With this notation, 
we have 
\begin{equation}
\myvec u = \myvec  u_{\mathrm{des}} + \mymatrix G \left(\tfrac{\partial \myvec  h}{\partial \myvec  q}\right)  \bm{\lambda}.
\end{equation}
} 
%
%
%
%
Finally, we define the state dependent feasible set under the linear complementarity constraints
\begin{multline}
    \mathcal{U}_{\mathrm{LC}} = \{ \myvec{u} = \myvec{u}_\mathrm{des}+ \mymatrix{G}\left(\tfrac{\partial \myvec{h}}{\partial\myvec{q}}\right)\bm{\lambda} \mid \bm{\lambda}\in \mathbb{R}^m,\\  
    \myvec{0} \leq \bm{\lambda} \perp \myvec{h} + \tau \tfrac{\partial \myvec{h}}{\partial\myvec{q}} \myvec{u} - \bm{\delta}_{\mathrm{LC}} \geq \myvec{0}
      \}. \label{eq:set_u_lcp}
\end{multline}

The optimization problem can be stated as
\begin{equation}
\label{eq:dtLCP}
\begin{aligned}
   \myvec{u}_{\mathrm{LC},k}^* 
   &= \arg\min_{\myvec{u}_k} \quad \|\,\myvec{u}_k - \myvec{u}_{\text{des}}(\myvec{q}_k)\|^2 
   \\
   & \quad \text{subject to} \quad 
       \myvec{u}_k \in \mathcal{U}_{\mathrm{LC}}^{(k)}(\myvec{q}_k),
\end{aligned}
\end{equation}
where $\myvec{u}_k$ is a function of $\bm{\lambda}_k$.
Note that the general form of the set \eqref{eq:set_u_lcp} can be stated as
\begin{multline}
    \mathcal{X}_{\mathrm{LC}} = \{ \myvec{x} = \mymatrix{G}\left(\mymatrix{A}_\mathrm{LC}\right)\bm{\lambda} \mid \bm{\lambda}\in \mathbb{R}^m,\\  
    \myvec{0} \leq \bm{\lambda} \perp \mymatrix{A}_\mathrm{LC} \myvec{x} - \myvec{b}_\mathrm{LC} \geq \myvec{0}
      \}, \label{eq:set_u_lcp_gen}
\end{multline}
with $\mymatrix{A}_\mathrm{LC}=\tfrac{\partial \myvec{h}}{\partial\myvec{q}}$, $\myvec{b_\mathrm{LC}} =-\tfrac{1}{\tau}(\myvec{h} - \bm{\delta}_\mathrm{LC}) -  \tfrac{\partial \myvec{h}}{\partial\myvec{q}} \myvec{u}_\mathrm{des}$, and $\myvec{x} = \myvec{u} - \myvec{u}_\mathrm{des}$, recovering the original set.
\subsection{CBF-based Approach}

A related approach for safe control has been developed through control barrier functions (CBFs) and QP-based control~\cite{ames2016control, ames2019control}, and also through vector field inequality approaches~\cite{marinho2018active}.
In this section we summarize relevant results from~\cite{breeden2021control} for CBFs in sampled-data or discrete systems.
An important result in CBF theory for control affine systems is that any Lipschitz continuous control input $\myvec{u}(\myvec{q})$ satisfying
\begin{align}
    \tfrac{\partial}{\partial \myvec{q}} h_i(\myvec{q}) \myvec{u} + \alpha_i (h_i(\myvec{q})) \geq 0,\quad i {=} 1,\dots,m, \quad \forall t{\geq} 0,
\end{align}
where $\alpha_i$ is a class-$\mathcal{K}$ function, renders $\Omega$ forward invariant~\cite{ames2019control, breeden2021control}.
A forward invariant set in the CBF sense describes that closed loop trajectories that start within the set, also remain within the set at all future times. 

For discrete or sampled-data systems, consider ZOH control laws with time step $\tau$
\begin{align}
    \myvec{u}(t) = \myvec{u}_k,\ \forall t \in [t_k,t_{k+1}),\ t_{k+1} = t_k + \tau.
\end{align}
The ZOH input trajectories satisfying
\begin{align}
    \tfrac{\partial}{\partial \myvec{q}} h_i(\myvec{q}_k) \,\myvec{u}_k + \alpha_i (h_i(\myvec{q}_k) -  \delta_{\mathrm{CBF},i})\geq 0, \label{eq:sampled-cbf}\\ i = 1,\dots,m,\quad k=0,1,\dots, \notag
\end{align}
at the sampled states $\myvec{q}_k$, are ensured to render $\Omega$ forward invariant.
Constraint~\eqref{eq:sampled-cbf} uses the physical margin formulation from~\cite{breeden2021control}, i.e., the variables $\delta_{\mathrm{CBF},i}$ are designed as a function of the global Lipschitz constants of the sampled dynamics and guarantee the safety of the underlying continuous system.


The feasible set of $\myvec{u}_k$ under these constraints is defined as
\begin{equation}
\label{eq:Ucbf}
\begin{aligned}
   \mathcal{U}_{\text{CBF}}(\myvec{q}) 
   := 
   \bigl\{
   & \myvec{u} \in \mathbb{R}^n \mid i=1,\dots,m,
   \\
   & \tfrac{\partial h_i}{\partial \myvec{q}}  \myvec{u} + \alpha_i (h_i -  \delta_{\mathrm{CBF},i})\geq 0
   \bigr\}.
\end{aligned}
\end{equation}
This set is convex, thus one can embed \eqref{eq:Ucbf} into the QP,
\begin{equation}
\label{eq:dtcbf_QP}
\begin{aligned}
   \myvec{u}_{\mathrm{CBF},k}^*
   &= \arg\min_{\myvec{u}_k} \quad \|\,\myvec{u}_k - \myvec{u}_{\text{des}}(\myvec{q}_k)\|^2 
   \\
   & \quad \text{subject to} \quad 
       \myvec{u}_k \in \mathcal{U}_{\text{CBF}}^{(k)}(\myvec{q}_k).
\end{aligned}
\end{equation}
This QP (CBF-QP) ensures that $\myvec{u}_k$ remains feasible w.r.t.\ 
the $m$ constraints $h_i(\cdot)$.

Note that the general form of the set \eqref{eq:Ucbf} can be stated as
\begin{align}
    \mathcal{X}_{\mathrm{CBF}} = \{ \myvec{x} \in \mathbb{R}^n \mid \mymatrix{A}_\mathrm{CBF} \myvec{x} - \myvec{b}_\mathrm{CBF} \geq \myvec{0}
      \}, \label{eq:set_u_cbf_gen}
\end{align}
with $\mymatrix{A}_\mathrm{CBF}=\tfrac{\partial \myvec{h}}{\partial \myvec{q}}$, $\myvec{b}_\mathrm{CBF} = -\bm{\alpha}  (\myvec{h} - \bm{\delta}_\mathrm{CBF}) - \tfrac{\partial \myvec{h}}{\partial \myvec{q}} \myvec{u}_\mathrm{des}$, and $\myvec{x} = \myvec{u} - \myvec{u}_\mathrm{des}$, recovering the original set.

Recall that for the first-order system $\dot{\myvec q}=\myvec u$, a standard CBF condition is $\dot{\bar h}_i(\myvec q)+\alpha_i(\bar h_i(\myvec q))\ge 0$, with $\alpha_i(\cdot)$ a class-$\mathcal{K}$ function. 
Since $\dot{\bar h}_i(\myvec q)=\frac{\partial \bar h_i}{\partial \myvec q}(\myvec q) \myvec u=\frac{\partial h_i}{\partial \myvec q}(\myvec q) \myvec u$, this yields the affine-in-input inequality $\frac{\partial h_i}{\partial \myvec q}(\myvec q_k)\myvec u_k\ge -\alpha_i \bigl(h_i(\myvec q_k)-\delta_i\bigr)$ at the sampling instants. 
On the other hand, the sampled-data one-step Euler enforcement used in the complementarity formulation can be written as $\myvec q_{k+1}=\myvec q_k+\tau\myvec u_k$, together with $h_i(\myvec q_{k+1})\gtrsim h_i(\myvec q_k)+\tau\frac{\partial h_i}{\partial \myvec q}(\myvec q_k)\myvec u_k$, and the clearance requirement $h_i(\myvec q_{k+1})\ge \delta_i$. 
Rearranging gives the discrete-time inequality $\frac{\partial h_i}{\partial \myvec q}(\myvec q_k) \myvec u_k\ge -\frac{1}{\tau}\bigl(h_i(\myvec q_k)-\delta_i\bigr)$. Therefore, for the common linear choice $\alpha_i(s)=k_i s$, selecting $k_i=1/\tau$ makes the continuous-time CBF inequality algebraically consistent with the one-step Euler condition at the sampling instants. This is exactly the identification exploited in our manuscript when matching the CBF-QP and LCQP constraints (cf. Eq. (41)), and it clarifies the distinct roles. The margin $\delta_i$ sets the enforced clearance level $h_i(\myvec q)\ge \delta_i$, while $\tau$ sets the one-step horizon and therefore the effective rate at which clearance must be recovered under sampled-data enforcement.

\section{Equivalence of Solutions} \label{sec:equivalence}

We show that the solutions $\myvec{u}_{\mathrm{LC}}^*$ and $\myvec{u}_{\mathrm{CBF}}^*$ to problems \eqref{eq:dtcbf_QP} and \eqref{eq:dtLCP} are equivalent by comparing the optimal solutions of the corresponding problems in general form.
First, we consider the single-constraint case, and then formalize and generalize the result to the multiple-constraint case in Theorem~\ref{lem:simple-proof}.

\subsection{Single Constraint Case}

\newcommand{\dq}{\dot{\mathbf{q}}}

Initially we consider the case of a single collision constraint for intuition, where we show the redundancy in the complementarity constraints under the minimum deviation objective, using a desired joint velocity $\dq_\mathrm{des}$.

The LCQP can be reformulated in terms of $\lambda$ as
\begin{align}
    \mathbf{u}^* &= \dq_\mathrm{des} + \mathbf{J}_c^\dagger \mathbf{n} \lambda^* \label{eq:single-lcqp-start} & \\
    \lambda^* &= \underset{\lambda}{\mathrm{argmin}} \ \lVert \mathbf{J}_c^\dagger \mathbf{n} \lambda \rVert^2 = \underset{\lambda}{\mathrm{argmin}} \ \lVert \lambda \rVert^2 \\
    & \quad \quad \quad \mathrm{s.t.} \quad  \dq = \dq_\mathrm{des} + \mathbf{J}_c^\dagger \mathbf{n} \lambda \\
    & \quad \quad \quad \quad \quad \ \, 0 \leq \lambda \ \perp \ h + \tau\mathbf{n}^T\mathbf{J}_c\dq \geq \delta, \label{eq:single-lcqp-end}
\end{align}
assuming $\lVert \mathbf{n}^T\mathbf{J}_c \rVert \neq 0$.
Expanding the complementarity constraint we obtain
\begin{align}
    0 \leq \lambda \ \perp \ h + \tau\mathbf{n}^T\mathbf{J}_c(\dq_\mathrm{des} + \mathbf{J}_c^\dagger \mathbf{n} \lambda) \geq \delta \\
    0 \leq \lambda \ \perp \ h + \tau\mathbf{n}^T\mathbf{J}_c \dq_\mathrm{des} + \tau\lambda \geq \delta,
\end{align}
which represents the three constraints,
\begin{align}
    0 \leq& \, \lambda  &\land \label{eq:left_c}\\
    \lambda(h + \tau\mathbf{n}^T\mathbf{J}_c \dq_\mathrm{des} + \tau\lambda - \delta)=& \, 0   &\land\label{eq:center_c}\\
    h + \tau\mathbf{n}^T\mathbf{J}_c \dq_\mathrm{des} + \tau\lambda - \delta \geq& \, 0  \label{eq:right_c}. &
\end{align}

To show the equivalence of problems, we show that the only limiting constraint is~\eqref{eq:right_c}.
Consider only the right inequality constraint \eqref{eq:right_c} in
\begin{align}
    \lambda^* = \underset{\lambda}{\mathrm{argmin}} \label{eq:single-obj2} \ & 
     \lVert \lambda \rVert^2 \\
    \mathrm{s.t.} \quad & h + \tau\mathbf{n}^T\mathbf{J}_c\dq_\mathrm{des} + \tau\lambda - \delta \geq 0, \label{eq:single-ineq2}
\end{align}
which results in a standard QP formulation.
Depending on the desired joint velocity, three distinct cases cover the possible solutions:
\begin{align}
    \mathrm{a.} & \quad h + \tau\mathbf{n}^T\mathbf{J}_c\dq_\mathrm{des} - \delta > 0, \implies \lambda^*=0, \\
    \mathrm{b.} &\quad h + \tau\mathbf{n}^T\mathbf{J}_c\dq_\mathrm{des} - \delta = 0, \implies \lambda^*=0, \\
    \mathrm{c.} &\quad h + \tau\mathbf{n}^T\mathbf{J}_c\dq_\mathrm{des} - \delta < 0, \notag \\ 
    & \quad \quad\implies \lambda^*=\frac{1}{\tau}(\delta - (h + \tau\mathbf{n}^T\mathbf{J}_c\dq_\mathrm{des})) > 0.
\end{align}
In all three cases, the optimal solution is non-negative, and therefore the inequality constraint \eqref{eq:left_c} is always satisfied.

In cases $(\mathrm{a.})$ and $(\mathrm{b}.)$ the equality constraint~\eqref{eq:center_c} is directly satisfied because of $\lambda^*=0$.
In case $(\mathrm{c}.)$ the optimal $\lambda$ is positive and leads to ${h + \tau\mathbf{n}^T\mathbf{J}_c\dq_\mathrm{des} + \tau\lambda^* = \delta}$, which in turn also satisfies \eqref{eq:center_c}.
Through the shown redundancy of the constraints \eqref{eq:left_c} and \eqref{eq:center_c}, we conclude that the optimal solutions to the problems \eqref{eq:single-lcqp-start}-\eqref{eq:single-lcqp-end} and \eqref{eq:single-obj2}-\eqref{eq:single-ineq2} are equivalent.

\subsection{General Case}

By considering the multiple-constraint case through the primitive form of the optimization problems, we show the equivalence of the optimal solutions in Theorem~\ref{lem:simple-proof}, which generalizes and formalizes the single-constraint analysis.

To map between the formulations, we first find the correspondence between variables and parameters between the approaches.
Since $\mymatrix{A}_\mathrm{CBF} = \mymatrix{A}_{\mathrm{LC}}$, and 
\begin{align}
    \myvec{b}_\mathrm{CBF} {=} 
 \myvec{b}_{\mathrm{LC}} \iff \bm{\alpha} (\myvec{h} - \bm{\delta}_\mathrm{CBF}) {=}  \tfrac{1}{\tau}(\myvec{h} - \bm{\delta}_\mathrm{LC}),
\end{align}
the approaches share the corresponding $\myvec{A}$ and $\myvec{b}$ variables when $\alpha_i(x) = \tau^{-1} x$, and when using the same margins $\bm{\delta} = \bm{\delta}_\mathrm{LC} = \bm{\delta}_\mathrm{CBF}$.

In the following theorem, the set $\mathcal{X}_0$ corresponds to the CBF-QP formulation, the set $\mathcal{X}_1$ corresponds to the LCQP formulation, and parameters are chosen such that $\mymatrix{A} = \mymatrix{A}_\mathrm{LC} = \mymatrix{A}_\mathrm{CBF}$ and $\myvec{b} = \myvec{b}_\mathrm{LC} = \myvec{b}_\mathrm{CBF}$.

\begin{theorem}[Equivalence of Solutions]
\label{lem:simple-proof}
Let $\myvec{b} \in \mathbb{R}^m$, and $\mymatrix{A} \in \mathbb{R}^{m \times n}$ with rows $\myvec{a}_i$ assuming $\lVert \myvec{a}_i\rVert\neq 0$. Using $\mymatrix{G}$ as in \eqref{eq:g-operator}, define
\begin{align}
    \mathcal{X}_0 &:=\{\myvec{x} \in \mathbb{R}^n \mid \mymatrix{A}\myvec{x} - \myvec{b} \geq \myvec{0}\} \quad\text{and} 
    \label{eq:x0convexset}
    \\
    \mathcal{X}_1 &:=\{\myvec{x} = \mymatrix{H} \bm{\lambda} \mid \bm{\lambda} \in \mathbb{R}^m,\ 
    0 \leq \bm{\lambda} \perp \mymatrix{A}\mymatrix{H} \bm{\lambda} - \myvec{b} \ge 0\},
    \label{eq:x1convexset}
\end{align}
where $\mymatrix{H} = \mymatrix{G}(\mymatrix{A})$.
Then the problems
\begin{align}
\min_{\myvec{x}} \tfrac12 \|\myvec{x}\|^2
\quad \text{subject to}\quad \myvec{x} \in \mathcal{X}_0
\quad\text{and} \label{eq:x0prob} \\
\min_{\myvec{x}} \tfrac12 \|\myvec{x}\|^2
\quad \text{subject to}\quad \myvec{x} \in \mathcal{X}_1 \label{eq:x1prob}
\end{align}
have the same optimal solutions. 
\end{theorem}

The formal proof below expands on the following argument. Since Problem~\eqref{eq:x0prob} is convex, the corresponding KKT conditions are necessary and sufficient for optimality. 
These enforce 
$\myvec{x} \in \mathrm{rowsp}(\mymatrix{A})$, and complementary slackness.  
\note{
Since 
$\operatorname{rowsp}(\mymatrix A)=\operatorname{colsp}(\mymatrix  H)$, that is, $\operatorname{rowsp}(\mymatrix  A)=\operatorname{rowsp}(\mymatrix H^T)$,} the minimizer of Problem~\eqref{eq:x0prob} is also a feasible and optimal solution to Problem~\eqref{eq:x1prob}. 
For a standard reference, see~\cite{boyd2004convex}.
\medskip

\noindent \textit{Proof.}
\note{We first show set inclusion.} 
If $\myvec{x} {\, \in \, } \mathcal{X}_1$, then by definition there exists \note{$\bm{\lambda} \in \mathbb{R}^m$ such that } 
\[
\myvec{x} = \mymatrix{H} \bm{\lambda}, 
\quad 
0 \le \bm{\lambda} \,\perp\, \mymatrix{A}\,\mymatrix{H} \bm{\lambda} - \myvec{b} \ge 0.
\]
Since $\mymatrix{A}\,\mymatrix{H} \bm{\lambda} {\ge} \myvec{b}$, it follows that 
\note{$\mymatrix A \myvec x {-} \myvec b =  \mymatrix A \mymatrix  H \bm \lambda {-}  \myvec b \ge 0$, and consequently,  }
$\myvec{x} \in  \mathcal{X}_0$, therefore $\mathcal{X}_1 \subseteq \mathcal{X}_0$.

\note{
Assume $\mathcal{X}_0$ is nonempty (otherwise $\mathcal{X}_1$ is also empty since $\mathcal{X}_1 \subseteq \mathcal{X}_0$, and both problems are infeasible).  
Problem~\eqref{eq:x0prob}  is the Euclidean projection of the origin onto the convex polyhedron $\mathcal X_0$.
Since the objective is strictly convex, and 
}
problem~\eqref{eq:x0prob} is a convex quadratic minimization with linear constraints, 
\note{it admits a unique minimizer, with }
$\myvec{x}_0^*$ being the optimal solution to Problem~\eqref{eq:x0prob}. 
\note{Moreover, the first-order optimality conditions for this strictly convex QP with affine inequalities yield the existence of multipliers  $\bm \lambda^\star \in \mathbb{R}^m$, $\bm{\lambda}^* \ge 0$}, satisfying 
Karush--Kuhn--Tucker (KKT) conditions, such that 
(Primal feasibility) ${\myvec{A}\,\myvec{x}_0^* \ge \myvec{b}}$, 
(Dual feasibility) $\bm{\lambda}^* \ge 0$, 
(Complementary slackness) $\bm \lambda_i^*\,( (\mymatrix{A}\,\myvec{x}_0^*)_i - \myvec b_i ) = 0 \ \forall \, i$, and 
(Stationarity) $\myvec{x}_0^* - \mymatrix{A}^T \bm{\lambda}^* = 0$.  
\note{In particular, stationarity gives $\myvec x_0^\star = \mymatrix A^T \bm \lambda^\star$, hence $\myvec x_0^\star \in \operatorname{rowsp}(\mymatrix A)$.}

\note{
We now connect this KKT multiplier 
$\bm{\lambda}^{\star}$ 
to the complementarity representation in $\mathcal X_1$. 
By construction, $\mymatrix{H}=\mymatrix{G}(\mymatrix{A})$ is built column-wise from the rows of $\mymatrix{A}$ via a nonzero row-wise scaling, which implies 
$\operatorname{colsp}(\mymatrix{H})=\operatorname{rowsp}(\mymatrix{A})$,  
under the standing assumption $\|\myvec{a}_i\|\neq 0$. 
Consequently, $\mymatrix{H}$ is surjective onto $\operatorname{rowsp}(\mymatrix{A})$. 
Since $\myvec x_0^{\star}\in \operatorname{rowsp}(\mymatrix{A})$,  
we also have that 
$\mymatrix{A}^{T}\bm{\lambda}^{\star}$ lies in the row space of $\mymatrix{A}$, i.e., in
$\operatorname{Im}(\mymatrix{A}^{T})=\operatorname{rowsp}(\mymatrix{A})$. 
%
From the definition $\myvec{a}_i^{\dagger}=(\myvec{a}_i)^{T}/\|\myvec{a}_i \|^2$, the columns of $\mymatrix{H}$ are nonzero scaled versions of the constraint normals. Therefore the image (column space) of $\mymatrix{H}$ coincides with the span of the rows of $\mymatrix{A}$, namely 
$\operatorname{Im}(\mymatrix{H})=\operatorname{colsp}(\mymatrix{H})=\operatorname{rowsp}(\mymatrix{A})=\operatorname{Im}(\mymatrix{A}^{T})$.
}

\note{
Hence there exists $\bm{\lambda}'$ such that $x_0^{\star}=\mymatrix{A}^{T}\bm{\lambda}^{\star}=\mymatrix{H}\bm{\lambda}'$. In particular, choosing
$\bm{\lambda}'_i=\|\myvec{a}_i\|^2\,\bm{\lambda}^{\star}_i$
yields 
$x_0^{\star}=\mymatrix{H}\bm{\lambda}'$. Thus,  
%
}
\[
\myvec{x}_0^* = \mymatrix{H} \bm{\lambda}', 
\quad 
0 \le \bm{\lambda}' \,\perp\, \mymatrix{A}\,\myvec{H} \bm{\lambda}' - \myvec{b} \geq 0,
\]
and any optimal solution $\myvec{x}_0^*$ of~\eqref{eq:x0prob} lies in $\mathcal{X}_1$. 
Since $\mathcal{X}_1 \subseteq \mathcal{X}_0$, it also follows that
the minimum values over $\mathcal{X}_0$ and $\mathcal{X}_1$ must coincide, i.e., problems \eqref{eq:x0prob} and \eqref{eq:x1prob} have the same optimal solutions. \qedsymbol
\medskip

As a consequence of theorem \ref{lem:simple-proof}, we can formally generalize the uniqueness results of complementarity-based safe control frameworks~\cite{sinha2021task}.

\begin{corollary}[Convexity of Safe Control]
Since $\mymatrix{A}_\mathrm{LC}\mymatrix{G}(\mymatrix{A}_\mathrm{LC})$ is positive semi-definite, the linear complementarity problem describing collision avoidance of robotic manipulators with velocity control, i.e., $\mathrm{find} \ \myvec{u}(\bm{\lambda})\in \mathcal{U}_\mathrm{LC}$, is convex.
Further, for every state $\myvec q$ and desired input $\myvec{u}_\mathrm{des}$, the set 
$\mathcal{U}_\mathrm{LC}$ is convex~\cite{den1993linear}.
\end{corollary}

\subsection{Geometric Lens on the Equivalence}

A helpful perspective for  understanding the equivalence between methods is that both the linear-complementarity via LCQP and CBF-based controllers are subject to constraints of the form   
${\myvec{A}\myvec{x} \geq \myvec{b}}$.  
These constraints can be viewed as 
${\bigcap_i \{ \myvec{x} : \myvec{a}_i\myvec{x} \geq b_i \}}$, where each 
$\myvec{a}_i$ depicts the normal to a constraint boundary--such as the gradient of a distance function. The feasible set corresponding to this linear constraint thus becomes a polyhedron or polyhedral cone formed by intersecting the half-spaces $\myvec{x} : \myvec{a}_i\myvec{x} \geq b_i$.

Demonstrating that a candidate control satisfies these constraints under both frameworks is central to establishing equivalence. This is achieved through the mapping $\mymatrix{H}$.  
Moreover, when solving Theorem \ref{lem:simple-proof} via \eqref{eq:x0prob} or \eqref{eq:x1prob}---under \eqref{eq:x0convexset} and \eqref{eq:x1convexset}, respectively---the minimal-norm solution $\myvec{x}^\ast$  must be orthogonal to every active facet of the polyhedron, i.e., $\myvec{a}_i\myvec{x}^\ast  \geq b_i$. If a facet is active, the corresponding multiplier $\lambda_i^\ast $ is strictly positive and represents the push needed to keep $\myvec x ^\ast  $ in the set. Otherwise, $\lambda_i$ is null. In both the complementarity and CBF-based approaches, these multipliers arise either through linear complementarity conditions or through KKT stationarity and slackness. As shown in the proof of Theorem \ref{lem:simple-proof}, these conditions align perfectly, yielding the same orthogonal projection-mapped through $\mymatrix H$ onto the same intersection of half-spaces---hence the same optimal solutions.

\subsection{Discussion}

While we demonstrate the equivalence between the considered safe whole-body control approaches for velocity-controlled manipulators, we argue that this benefits both directions and motivates future research. 
Specifically, establishing a direct correspondence between the two frameworks enables the transfer of algorithmic improvements 
and theoretical guarantees.

We pose that CBF methods benefit from the connections to a larger set of problems closer to contact applications and using complementarity constraints.
Further, planning applications as in \cite{chakraborty2009complementarity, sinha2021task} could be useful guides for the design of planning methods based on CBFs.
On the other hand, the well-developed CBF theory is helpful for discrete complementarity approaches, e.g., sampled-data CBFs also provide formal statements on valid physical margins and sampling times depending on the Lipschitz constants of the systems and constraints~\cite{breeden2021control}, ensuring the safe behavior of complementarity-based approaches.
Extending the analysis to higher order dynamics and other more complex systems could lead to further connections between the two methods. 

\note{Finally, it is also worth highlighting that, from a theoretical viewpoint, Theorem 1 shows that both approaches represent the same minimum-deviation safety projection step. In both cases, one solves a convex QP that minimally corrects a nominal input subject to the same set of affine, linearized safety inequalities, so the two methods share the same polyhedral feasible set with the same constraint normals and the same optimality conditions. In this sense, the CBF formulation enforces the half-space constraints directly through barrier inequalities at each sample, while the complementarity formulation makes the active or inactive facet logic explicit through multipliers associated with the same half-spaces. Consequently, geometric sources of ill-conditioning in constraint-based solutions, including poor scaling, near-parallel or degenerate active facets, and tolerance sensitivity near activation, are expected to be shared across both viewpoints. In practice, however, numerical behaviour can still differ due to solver choice such as active-set, interior point methods, or augmented Lagrangian methods, and due to implementation details including stopping criteria, regularization, scaling, and warm-starting. Interestingly, this leaves room for designers to select and tune the numerical pipeline best suited to their application, while still preserving the same theoretical structure and the equivalence guarantees under the modeling assumptions. Notwithstanding, a detailed numerical conditioning study is beyond the scope of this work, which is aimed at establishing theoretical guarantees rather than investigating empirical performance and engineering trade-offs.
}
\section{Numerical Validation}
\label{sec:validation}

We validate the obtained equivalence result via a numerical experiment on a three-degree-of-freedom (3-DoF) planar serial-chain robot in a simulated environment with an obstacle. 
\note{The objective of this numerical example is not to demonstrate superiority in large-scale or highly cluttered environments, but rather to provide a controlled and transparent validation of the theoretical claims. For this reason, we deliberately adopted a simplified 3-DOF planar setup with a single obstacle. This choice allows the equivalence to be isolated and clearly illustrated without additional confounding factors introduced by complex geometry, high-dimensional kinematics, or implementation-specific tuning. Importantly, the equivalence result itself is independent of the specific obstacle geometry or system dimension, as it follows directly from the structural properties of the underlying inequality constraints. Notwithstanding, we refer to instances where LCQP~\cite{yao2025synthesis,huang2026unified,sinha2021task,sinha2023oc3} and CBF-QP~\cite{murtaza2021real,murtaza2022consensus,murtaza2022safety} have already been implemented and validated in substantially more complex scenarios in prior literature, including multi-obstacle environments and higher-DOF systems.} Figure~\ref{fig:validation} shows the environment and an example robot configuration along the collision-free path.

\subsection{Setup}

The system is simulated with a time step of $\tau = 5\,\mathrm{ms}$. The link lengths are $\mathbf{l} = (0.1,\, 0.05,\, 0.05)$. A single disk-shaped obstacle of radius $0.05$ is placed at $(0.03,\, 0.170)^T$, and the physical safety margin is $\delta = 0.01$. The robot base is positioned at $\mathbf{p}_\mathrm{base} = \mathbf{0}$, and the goal location is $\mathbf{p}_g = (-0.05,\, 0.15)^T$. The end-effector position is defined as $\mathbf{p}_\mathrm{ee} = \mathbf{f}_\mathrm{ee}(\mathbf{q})$, and the desired joint velocity is given by
\begin{align}
    \dot{\mathbf{q}}_\mathrm{des} = \big(\tfrac{\partial \myvec{f}_\mathrm{ee}}{\partial \mathbf{q}}\big)^\dagger \mathbf{v}_\mathrm{des} \\
    \mathbf{v}_\mathrm{des} = k_p(\mathbf{p}_g - \mathbf{p}_\mathrm{ee})\lVert \mathbf{p}_g - \mathbf{p}_\mathrm{ee} \rVert^{-1}.
\end{align}

We use \texttt{quadprog} to solve the CBF-QP formulation and \texttt{fmincon} to solve the complementarity problem in \textsc{Matlab} (2024b), relying on standard parameters and convergence tolerances of the solvers.

\begin{figure}
    \centering
    \includegraphics[width=\linewidth]{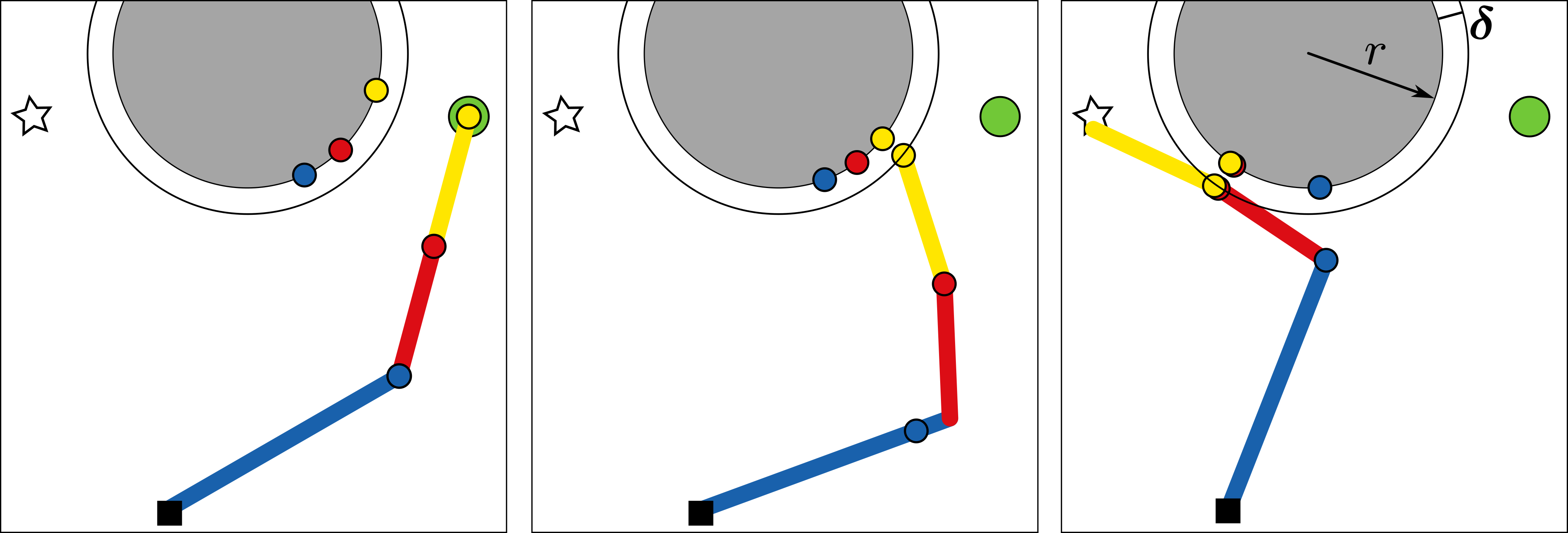}
    \caption{A 3-DoF planar robot is guided from an initial configuration (left) to reach a goal with the end effector, depicted here through the star. The robot follows the complementarity and the QP-CBF policies, leading to identical paths. The illustration in the center shows an example configuration along the path, and the one on the right shows the robot configuration when reaching the goal. The notation described in the right figure applies equally to the left and center figures.}
    \label{fig:validation}
\end{figure}

Since the obstacle is a single disk, three scalar constraints are introduced, each enforcing collision avoidance between one robot link and the obstacle. In Figure~\ref{fig:validation}, the pairs of closest points on each link and on the obstacle surface are shown as filled circles sharing consistent colors.

\begin{figure}
    \centering
    \includegraphics[width=\linewidth]{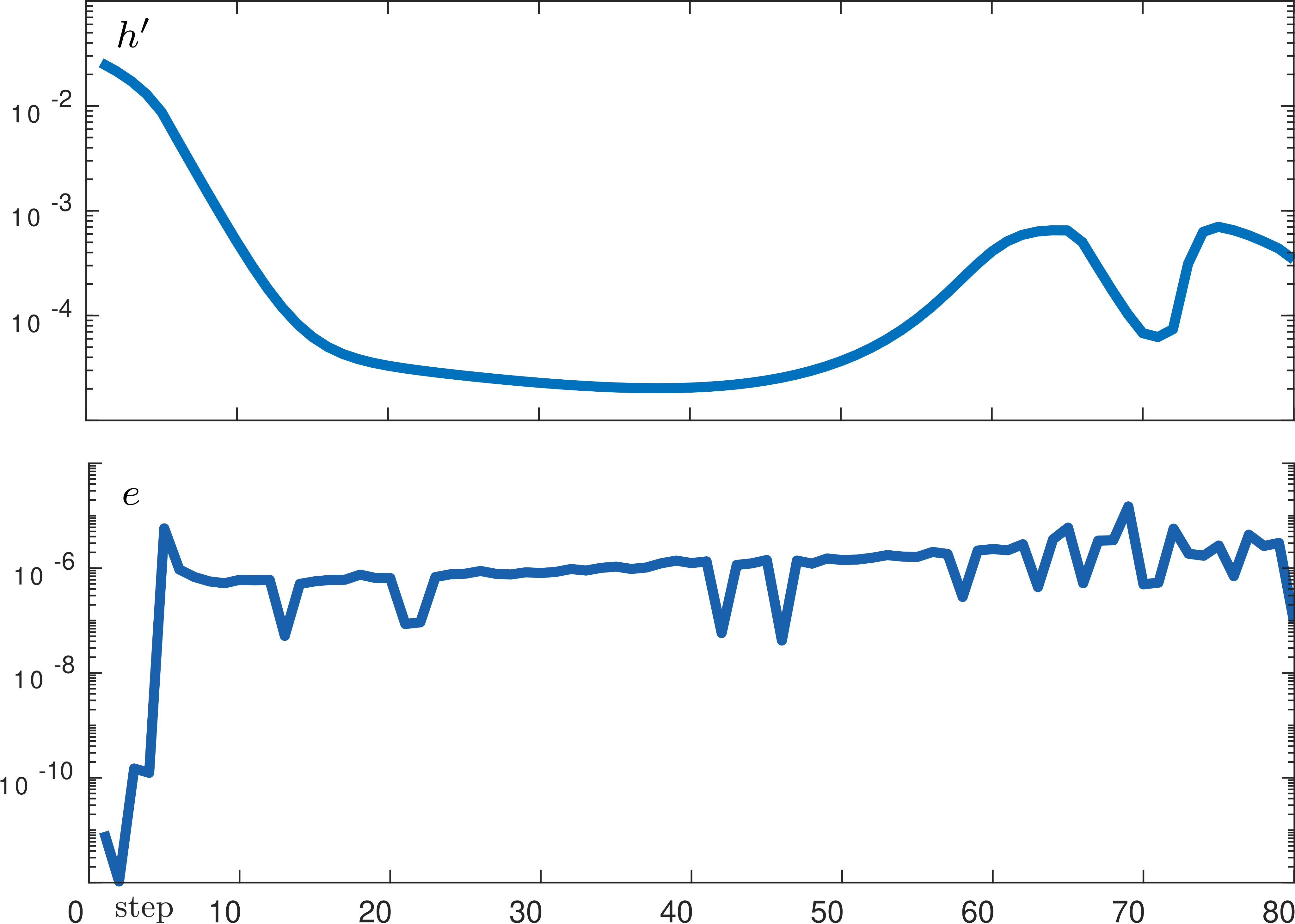}
    \caption{The safety constraint $h'\geq0$ (top) and the solution error $e$ (bottom) are plotted at each time step of the simulation.}
    \label{fig:constraint-error}
\end{figure}

\subsection{Results}
We define the solution error at time step $k$ as $e_k = \lVert \mathbf{u}_{\mathrm{LC},k} - \mathbf{u}_{\mathrm{CBF},k} \rVert$.
The solution error at each time step is shown in Figure~\ref{fig:constraint-error}; error statistics over the entire trajectory are
\begin{align}
    [\min, \mathrm{mean}, \max](e) = [1.1\text{e-}12,\, 1.5 \text{e-}6,\, 1.5\text{e-}6].
\end{align} 
These values demonstrate that the solutions computed by the two methods match up to the numerical solver tolerances, supporting the claimed equivalence. 

The value of the reduced constraint, $h' = \min(\mathbf{h}) - \delta$, remains positive throughout the motion, with $\min(h') = 2.0\times10^{-5}$.
The reduced constraint at each time step is shown in Figure~\ref{fig:constraint-error}. 
This indicates that the resulting trajectory is collision-free and respects the imposed physical margin.

\section{Conclusion} \label{sec:conclusion}

We describe and compare CBF-based methods and complementarity-based approaches for safe and reactive whole-body robot control in the case of discrete time first order dynamics.
Using redundancy arguments we show the equivalence between optimal solutions in the single-constraint case. With KKT arguments from convex optimization, we prove the equivalence of the optimal solutions in the general multiple-constraint case.
We pose that the equivalence result provides both practical and theoretical benefits to the individual methods, which are usually utilized separately.
These can lead to further tools and insights, effectively motivating future work in the equivalence analysis for more general systems and constraints.

\bibliographystyle{IEEEtran}
\balance
\bibliography{ref}

\end{document}